\documentclass[graybox]{svmult}
\usepackage{gensymb}
\usepackage{graphicx}
\usepackage{float}
\usepackage{xfrac} 
\usepackage{fancyhdr}
\usepackage{epstopdf}
\usepackage{amsmath}

\usepackage{amssymb}
\usepackage{algorithm}
\usepackage{algpseudocode}
\usepackage{pifont}
\usepackage[caption=false]{subfig}
\usepackage[numbers]{natbib}
\usepackage{graphicx}
\usepackage{graphbox}

\usepackage[utf8]{inputenc}

\usepackage{geometry}
\begin{document}
\title*{Multisensory Omni-directional Long-term Place Recognition: Benchmark Dataset and Analysis}
\author{Ashwin Mathur, Fei Han, and Hao Zhang}

 \titlerunning{Multisensory Long-term Place Recognition} 

\institute{\email{mathurash2009@gmail.com, fhan@mines.edu, hzhang@mines.edu}
\and \at Colorado School Of Mines, Golden, CO 80401, USA.}

\maketitle

\abstract
\\{Recognizing a previously visited place, also known as place recognition (or loop closure detection) is the key towards fully autonomous mobile robots and self-driving vehicle navigation.
Augmented with various Simultaneous Localization and Mapping techniques (SLAM), loop closure detection allows for incremental pose correction and can bolster efficient and accurate map creation.
However, repeated and similar scenes (perceptual aliasing) and long term appearance changes (e.g. weather variations) are major challenges for current place recognition algorithms. 
We introduce a new dataset \emph{Multisensory Omnidirectional Long-term Place recognition} (MOLP) comprising omnidirectional intensity and disparity images. This dataset presents many of the challenges faced by outdoor mobile robots and current place recognition algorithms. Using MOLP dataset, we formulate the place recognition problem as a regularized sparse convex optimization problem. We conclude that information extracted from intensity image is superior to disparity image in isolating discriminative features for successful long term place recognition. Furthermore, when these discriminative features are extracted from an omnidirectional vision sensor, a robust bidirectional loop closure detection approach is established, allowing mobile robots to close the loop, regardless of the difference in the direction when revisiting a place.

\section{Introduction}
\label{sec:Intro}
The ability to map and localize in that map, also known as Simultaneous Localization and Mapping (SLAM), is a major hurdle for autonomous robot navigation. 
In the last two decades a variety of solutions to the SLAM problem for both indoor and outdoor mobile robots have been published \cite{PR,FABRAT,RVR,RLC,HPLC,BG}. 
Today, many applications require robots to function autonomously in an environment over a long period of time (months or years), for example, in self-driving applications. 
This is a challenging problem, as environments change due to weather and seasonal patterns as well as dynamic objects such as people and automobiles hinders the robots ability to learn and recognize places accurately.

Many studies in the robotics community identified place recognition \cite{FIM,FABRAT,RVR,RLC,HPLC,BG} as an essential ability for robots to localize in a dynamic environment.
Place recognition is the ability of a mobile robot to localize in a given environment by recognizing previously visited places, also known as loop closure detection. Other than localization, the ability for mobile robots to close the loop \cite{RVR,RLC,HPLC,BG} is highly desirable as this can be used to correct incremental pose drift due to motion and sensor errors. As a consequence of this correction,
accurate maps can be created \cite{OGLC}.

Closing the loop is challenging for two reasons: perceptual aliasing problem \cite{zhang2016robust} and long-term appearance changes \cite{SRAL}. Perceptual aliasing is the apparent similarity between two different locations, which is a common trait of indoor (e.g., corridors) and outdoor (e.g., highways) environments. 
Whereas, long-term appearance changes are the variations in appearance of a location due to long term effects like seasonal and weather changes.

Current research efforts have focused on solving these issues by learning discriminative spatial and/or temporal patterns in an environment. 
Both LiDARs \cite{LOAM} and RGB-D cameras \cite{FABRAT,RVR,RLC,HPLC,BG} have been utilized to learn these patterns. 
Using the fact that loop closure events yield sparse solutions \cite{OSC}, 
promising results have been obtained through fusing multi-modal local and global features and enforcing sparsity using group-norms \cite{SRAL}.
Enforcing sparsity in this fashion yields discriminative features that describe a scene, 
resulting in a robust solution to the long-term appearance change problems. 
Few methods \cite{SEQ,RVR} have shown that matching a query image with many possible sequences and then finding a global match yields robust solutions to the perceptual aliasing and long term appearance change problem.

In this paper, we introduce a new large-scale dataset comprising sequences of omnidirectional intensity and stereo images
to provide a benchmark dataset to evaluate and compare methods for multisensory long-term place recognition under the challenges of perceptual aliasing and long-term appearance changes.
We have chosen to collect data in two distinct environments (city and mountain routes), four different seasons (spring, summer, fall and winter) and two different times (morning and evening) to capture various characteristics typically seen in an outdoor setting, long-term appearance changes and perceptual aliasing occurrences. Lastly, for each route the data was collected in two directions (i.e., going forward and backward along the road). 

To analyze this dataset, we implement a regularized convex optimization algorithm based on \cite{SRAL}.
The algorithm was developed on the idea that loop closure events yield sparse solutions \cite{SRAL}, therefore only a confined set of sensor modalities (e.g., intensity image) and feature modalities (e.g., HOG features) are required to describe an environment and recognize revisited places (i.e. loop closure detection).
These sensor and feature modalities can also be called discriminative attributes of an environment, that are impervious to disturbances from dynamic objects, weather and illuminations changes.
The algorithm learns the weights of these discriminative features and sensor modalities and places higher weights on corresponding features during image matching, resulting in long-term loop closure detection.

The major practical contributions of this paper are as follows:

\begin{itemize}
	\item Introduction of a new large-scale dataset comprising omnidirectional intensity and disparity sequences as a benchmark for \emph{Multisensory Omnidirectional Long-term Place recognition} (MOLP). Details of the MOLP benchmark dataset is available at \url{http://hcr.mines.edu/code/MOLP.html}.
	\item Practical analysis of the effectiveness of the disparity image sensory modality in conjunction with intensity image modality against long term appearance changes and perceptual aliasing challenges.
	\item  Analysis of how multimodal features from an omnidirectional camera can aid in loop closure detection, when learning and loop closure detection are done in opposing directions
\end{itemize}

The remainder of this paper is organized as follows. Section \ref{sec:Rel} discusses existing visual place recognition techniques. Details of our Omni-directional dataset is presented in Section \ref{sec:ODD}. The convex optimization algorithm is introduced in \ref{sec:MOLP}. Experimental results are delineated in Section \ref{sec:Exp}. Lastly, conclusion and future work are discussed in Section \ref{sec:con}.

\section{Related Work}
\label{sec:Rel}
Due to wide availability and low cost, most existing place recognition methods rely on vision as the primary sensory modality. Visual place recognition algorithms can be differentiated based on three attributes: the type of information from images that is being extracted, how that information is being stored and how that information is retrieved for recognizing previously visited places.

For reliable matching, features are extracted from images in the form of descriptor vectors, that are categorized as local or global features. In conjunction with a bag of words approach, SIFT has been utilized successfully many times for visual place recognition \cite{SIFT_1,FIM}. 
Recent methods like FAB-MAP \cite{FAB} used SURF \cite{SURF} and Chow Lui vocabulary tree method to infer revisited places. Although invariant to rotation and scaling, local features often fail during varying lighting and environmental conditions.
Global descriptors can be applied to the whole image or selected patches of the image. 
HOG \cite{HOG} is a popular global descriptor that stores the gradient variations in each pixel in a histogram,
which is used in \cite{HOG_2} to find scene signatures that robustly describe an environment despite varying lighting and weather conditions. Recently, convolutional neural networks \cite{CNN,CNN_1} have been used to extract robust features against long term appearance changes. 

Aside from visual sensors, many methods used geometric information of the environment using RGB-D sensors \cite{RGBD} for SLAM, which then augment place recognition techniques for updating the map.
Depth information has also been used for object recognition, which becomes the basis to recognize revisited places \cite{SLAM++}. 
This technique is efficient in indoor environments with repeated and identical objects, however fails with environments changing in the long term. Cadena et. al. used stereo cameras with a bag of words technique and condition random fields (CRF) to avoid any geometrically inconsistent matches \cite{ST} to detect loop closures. 
Furthermore, Cadena et. al. mentioned the flexibility of CRF-matching algorithms allowing them to fuse other feature modalities like color information for more robust loop closure detection. 
Fusing various sensors and feature modalities is advantageous as one modality can capture a specific scene signature that another modality cannot and vice versa. 
This yields a robust solution to the perceptual aliasing and long-term appearance variation issues. 
However, combining the information from various modalities is computationally expensive making implementation on mobile robots challenging.
Algorithms like shared representative appearance learning (SRAL) \cite{SRAL} learned the weights of features that are shared among various scenarios (e.g., spring, summer, fall, winter) and identified these as discriminative and important features during the matching process. 
This allows one to efficiently augment multiple feature and sensor modalities during the training phase and then only utilize the important modalities during the matching phase. Our algorithm is based on the same idea of isolating discriminative features and sensor modalities using sparsity inducing group norms \cite{SRAL}.

\section{Omni-directional Dataset}
\label{sec:ODD}

In order to provide a benchmark for multisensory omnidirectional long-term place recognition, 
we introduce the new, large-scale MOLP dataset that captures a range of challenges (including perceptual aliasing, long-term appearance changes, and environment dynamics)  that could be encountered in an outdoor setting.
To record the data, we used the Occam Vision Group's 3.2 megapixel Omni Stereo camera and a GPS module for ground truth. The setup is shown in Figure \ref{fig:setup}. 
The Omni Stereo has 10 individual synchronized and calibrated camera streams outputting at a resolution of $480 \times 752$ each. Stitching these image streams yields an omnidirectional intensity and disparity image with a 58$^{\circ}$ vertical field of view. A GPS unit is used for ground truth, which outputs Latitude and Longitude positions at 1 Hz. A simple linear interpolation is carried out to calculate GPS coordinates corresponding to all the images. The data were collected using Robot Operating System (ROS) packages on an Ubuntu 14.04 laptop (Intel i7 2.4 GHz CPU and 16 GB memory).

\begin{figure}
	\centering
	\includegraphics[width=4in]{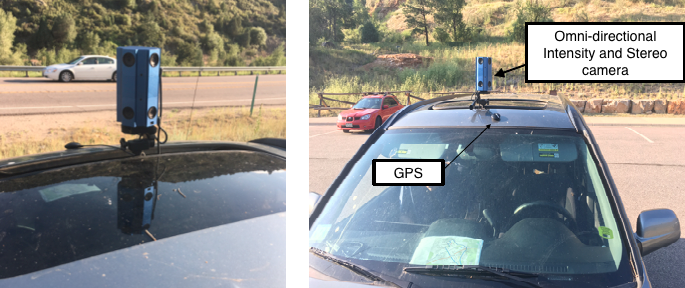}
	\caption{Camera and GPS setup. Camera settings (e.g. exposure level) were optimized at the start of each run in order to get the best intensity and stereo image.}
	\label{fig:setup}
\end{figure}

The dataset was collected on two distinct routes: \emph{city route} and \emph{mountain route},
as illustrated in Figure \ref{R_1} and \ref{R_2}.
The city route is approximately a 4.3 mile drive in the areas of the Golden downtown and the campus of Colorado School of Mines, which takes approximately 12-20 minutes to complete. 
Each run consists of 3000-7000 frames with a video resolution of $980 \times 3760$ (intensity + disparity image) and frame rate between 5-10 frames per second (FPS). 
The city route captures dynamic objects such as traffic, pedestrians and construction work.
The mountain route is a 10.3 mile drive and takes approximately 10-15 minutes to complete. 
Each run consists of 2500-5000 frames with a video resolution of $980 \times 3760$ (intensity + disparity image) and a frame rate between 5-10 FPS. 
The mountain route is a good example of a dataset with high perceptual aliasing due to the repetitive nature of the route. 
Furthermore, large illumination changes are seen in the mountain route dataset which is attributed to the route being surrounded by tall mountains and tunnels. Examples of images from each route can be observed in Figure \ref{example}. 

\begin{figure*}[t!]\label{fig:route}
\centering
\subfloat[City Route Map{\label{R_1}}]{\includegraphics[width=55 mm]{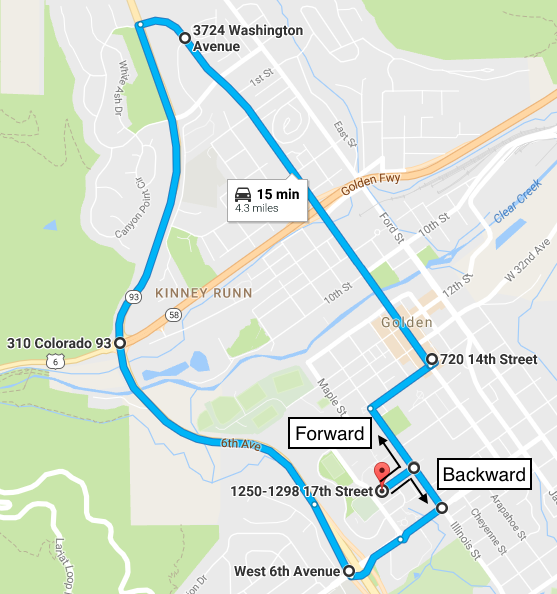}}
\quad
\subfloat[Example Dataset Frames{\label{example}}]{\includegraphics[width=55 mm]{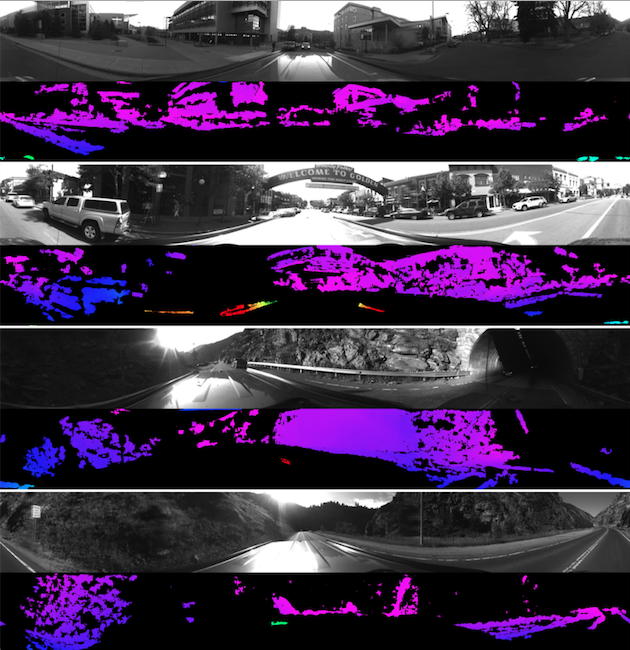}}
\quad
\subfloat[Mountain Route Map{\label{R_2}}]{\includegraphics[width=110 mm]{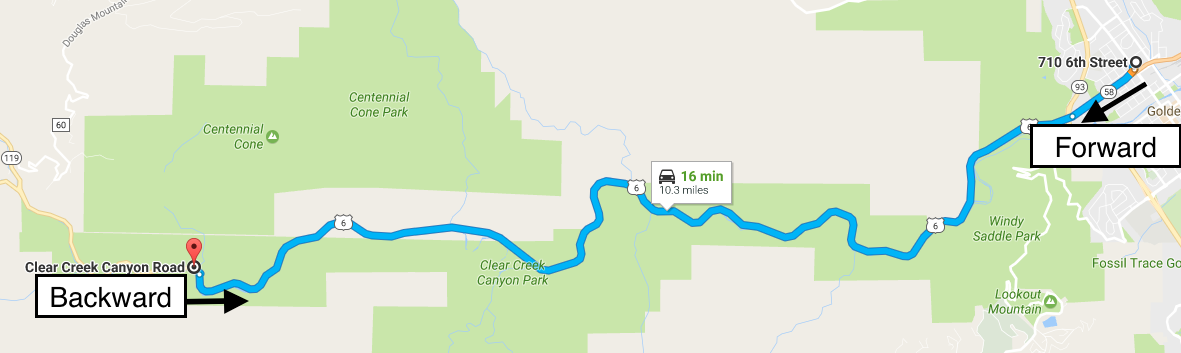}}
\caption{Driving Routes. Fig. \ref{R_1} illustrates the city route. City route was a loop driving in both directions as noted by the arrows. Fig. \ref{example} illustrates a set of two example images from route A and B on the top and bottom respectively. Fig. \ref{R_2} illustrates the mountain route. The mountain route forward direction was going up the mountain and backward direction was down the mountain. Map data \textcircled{c}2017 Google.}\label{fig:route}
\end{figure*}	

Each route was traversed in the mornings (about 9-11 am) and evenings (about 6-8 pm) multiple times to capture the illumination changes and dynamic activities like vehicle and pedestrian traffic. Also, datasets were collected in Spring (March-April), summer (July-August), fall (November) and winter(January-February), to capture the effects of long term changes due to illumination variations, weather, vegetation and construction activities. Lastly, data for each route in different times of day and seasons were collected, while driving in opposing directions to analyze the issue of bidirectional loop closure detection.

\section{Convex Optimization Algorithm}
\label{sec:MOLP}
Following the idea in \cite{SRAL},
we implement a sparse convex optimization method that learns the importance of the sensors (intensity and disparity) and multiple feature modalities to describe an environment in a discriminative fashion,
in order to practically analyze our newly collected MOLP dataset.

After data from different scenarios (e.g. summer, fall) are collected, various types of features from multiple sensory modalities (e.g., intensity and disparity) can be extracted from \textit{n} images and expressed as, $\mathbf{A} = [\mathbf{A_1;A_2;...;A_p}]\in\Re^{p\times n}$, where each sensor modality 
$\mathbf{A_i} = [\mathbf{a_1},...,\mathbf{a_n}] \in\Re^{d\times n}$, 
consists of $m$ feature modalities for each image, $\mathbf{a_i} = [\mathbf{(a_i^1)^\top},...,\mathbf{(a_i^m)^\top}]^\top \in\Re^{d}$ , 
where the total size of the descriptor vector from all sensory modality is, $p = \sum_{j=1}^p \sum_{i=1}^m{d_i}_j$, where ${d_i}_j$ is the descriptor vector size of the $i$-th feature modality from the $j$-th sensor.  
Each image used in the training has to have a specified scenario, $c$,  which is stored in $\mathbf{B}=[\mathbf{b_1},...,\mathbf{b_n}]^\top\in\Re^{n\times c}$, where ${b_i}_j\in\{ 0,1\}$, is representative of the $i$-th image existence in the $j$-th scenario. The optimization algorithm to calculate the discriminative features and sensor modalities can be formulated as an unconstrained regularized minimization problem:

\begin{equation} \label{eq:min}
	\min_{\mathbf{W}} \boldsymbol\ell{\mathbf{(W)}} + \lambda_1 \Omega{\mathbf{(W)}} ,
\end{equation}
where $\mathbf{W} \in\Re^{p\times c}$ is the weight matrix with $w_{ij}$ representing the importance of the $i$-th feature in reference to the $j$-th scenario. Picking a Frobenius loss function and $M$-norm feature grouping regularization term gives the following objective function:
\begin{equation} \label{eq:min1}
	\min_{\mathbf{W}} {\Vert \mathbf{A^\top W - B} \Vert}_F + \lambda_1 \Vert\mathbf{W}\Vert_M ,
\end{equation}

The Frobenius loss function is indicative of the residual error in describing an image in a scenario with the weighted multimodal features. $\lambda_1$ is the trade-off hyper parameter which signifies how much importance is given to the feature grouping norm compared to the loss function. To obtain discriminative feature modalities, which are required for robust loop closure detection, the $M$-norm or feature modality regularization term is chosen\cite{SRAL}. The $M$-norm promotes discriminative feature modalities, which are present in all scenarios:
\begin{equation} \label{eq:min1}
	\Vert\mathbf{W}\Vert_M = \sum_{q=1}^p \sum_{k=1}^m \sqrt{\sum_{j=1}^{d_i} \sum_{i=1}^c (w^q_{ijk})^2} = \sum_{q=1}^p \sum_{k=1}^m \Vert\mathbf{W^q_k}\Vert_F  ,
\end{equation}
where $\Vert\mathbf{W^q_k}\Vert_F \in\Re^{d_k\times c}$  is the weight matrix corresponding to the $k$-th feature modality from $q$-th sensor modality. 

Since two sensor modalities are present in our experiment, the underlying structure of the sensor modalities also needs to be incorporated to the problem formulation.
As an extension of \cite{SRAL},
we propose a new $S$-norm, which operates on all the feature modalities from a given sensor modality (i.e., intensity or disparity).
The final objective function with the $S$-norm is as follows:
\begin{equation} \label{eq:min2}
	\min_{W} {\Vert \mathbf{A^T W - B} \Vert}_F + \lambda_1 \Vert\mathbf{W}\Vert_M + \lambda_2 \Vert\mathbf{W}\Vert_S ,
\end{equation}
where $\lambda_2$ is the hyper-parameter associated with the $S$-norm and the $S$-norm is defined as:
\begin{equation} \label{eq:min3}
	\Vert\mathbf{W}\Vert_S =  \sum_{q=1}^p \sqrt{\sum_{k=1}^m \sum_{j=1}^{d_i} \sum_{i=1}^c (w^q_{ijk})^2} = \sum_{q=1}^p \Vert\mathbf{W^q}\Vert ,
\end{equation}
where $\Vert\mathbf{W^q}\Vert \in\Re^{d_q\times c}$ is the weight matrix corresponding to the $q$-th sensor modality.

The $S$-norm promotes the weights of sensor modalities that are representative in all the scenarios, i.e., discriminative sensor modalities are assigned a high weight, otherwise low weights (close to 0) are assigned to sensor modalities that are only representative in a subset of all the scenarios. The combination of $M$-norm and $S$-norm yields the desired sparse solution and lead to a small set of features and sensor modalities that allows for loop closure detection in the presence of dynamic objects, occlusions, illumination changes, long term appearance changes, and perceptual aliasing challenges. Lastly, due to the convex nature of the objective function, any optima will be a global optima.

After the training phase is completed \cite{SRAL}, query images can be matched to a bag of images collected during the training phase using a similarity score. Features from all sensor modalities need to be extracted from the query image and a similarity score is calculated using the following equation:
\begin{equation} \label{eq:min3}
	s =  \sum_{q=1}^p \sum_{i=1}^m {\bar{w}^q}{\widetilde{w}^ q_i} {s^q_i} ,
\end{equation}
where $s^q_i$ is a distance norm between the query and stored image's  $i$-th feature in $q$-th sensor modality, ${\widetilde{w}^ q_i} = \Vert\mathbf{W^q_i}^*\Vert_F$ is the optimal weight of the $i$-th feature in $q$-th sensor modality and ${\bar{w}^q} = \Vert\mathbf{W^q}\Vert_F$ is the optimal weight of the $q$-th sensor modality. If the two images being compared are of the same place then this score will be close to one. A predefined threshold can be used to conclude if the query image is a match to a previously visited location.

\section{Experimental Results}
\label{sec:Exp}

The results phase was divided into three parts: feature extraction, training and testing (matching). 
No post processing of either the intensity or disparity images have been conducted. 
During the feature extraction phase, GIST\cite{GIST}, HOG\cite{HOG}, LBP\cite{LBP} and CNN\cite{CNN} feature modalities are extracted from each sensor modality (intensity and disparity images). 
All features are extracted on a down sampled image size of $120 \times 752$ for both the disparity and intensity image. Due to the lack of good winter data during the training phase, only summer morning, summer evening and fall evening scenarios are taken into consideration during the training phase. The hyper parameters $\lambda_1$ and $\lambda_2$ are set to 0.1 and 0.01 respectively, throughout the training phase. To avoid overfitting, different training and testing datasets are used for each experiment. Two images are considered to be of the same place if they are separated by less than approximately 50 m. Lastly, C++ and Matlab programs are utilized during the training and testing phase, on an Intel i7 3.5GHz CPU, 16GB RAM with GeForce TITAN GPU.

\begin{figure*}[t!]
\hspace*{\fill}%
\centering
\subfloat[Scene Matches{\label{C_D1}}]{\includegraphics[width=4.6in,height=1.5in]{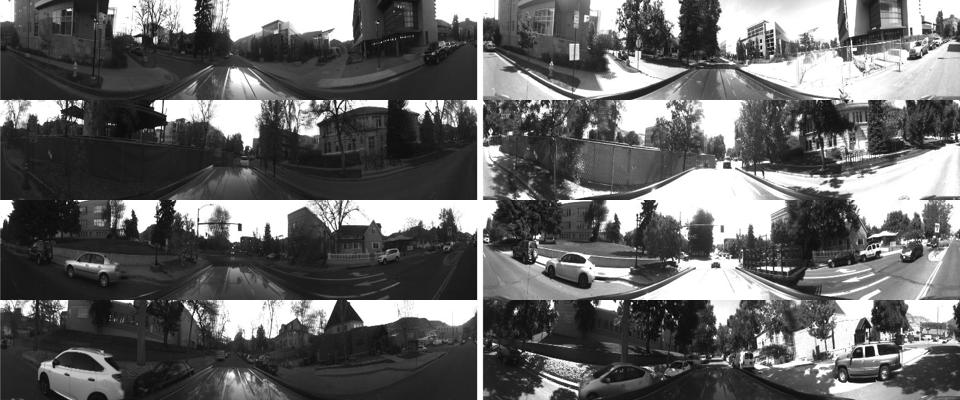}}%
\hfill
\subfloat[Precision-Recall Curves{\label{C_D2}}]{\includegraphics[width=.5\textwidth]{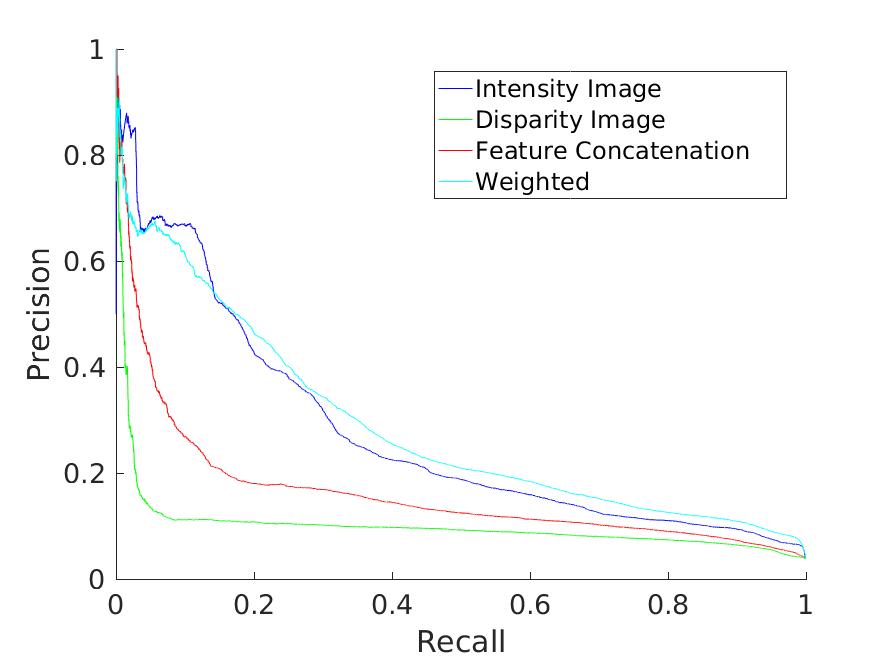}}%
\hfill
\subfloat[Modality Importance Weights{\label{C_D3}}]{\includegraphics[width=.5\textwidth]{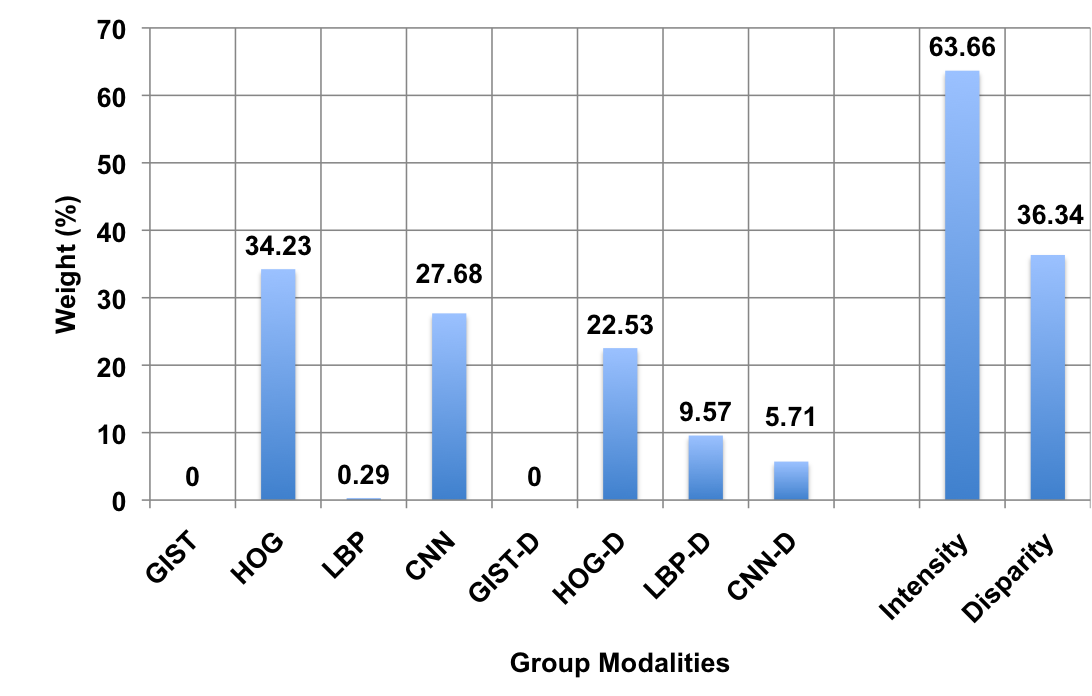}}%
\label{fig:weight_city}
\hfill
\caption{City route -- intensity versus disparity data. Fig. \ref{C_D1} presents Fall evening and Summer morning scene matches on the left and right side respectively. These matches were chosen to delineate long term appearance changes, illumination changes and dynamic activity. Fig. \ref{C_D2} illustrates a performance comparison using precision-recall curves. The weighted line represents the results from our optimization algorithm. Fig. \ref{C_D3} illustrates modality importance weights. Each feature modality was extracted from each sensor modality. Feature modalities from disparity images have an additional '-D' at the end.}\label{fig:cityd}
\end{figure*}	

\begin{figure*}[t!]
\centering
\hspace*{\fill}%
\subfloat[Scene Matches{\label{M_D1}}]{\includegraphics[width=4.6in,height=1.5in]{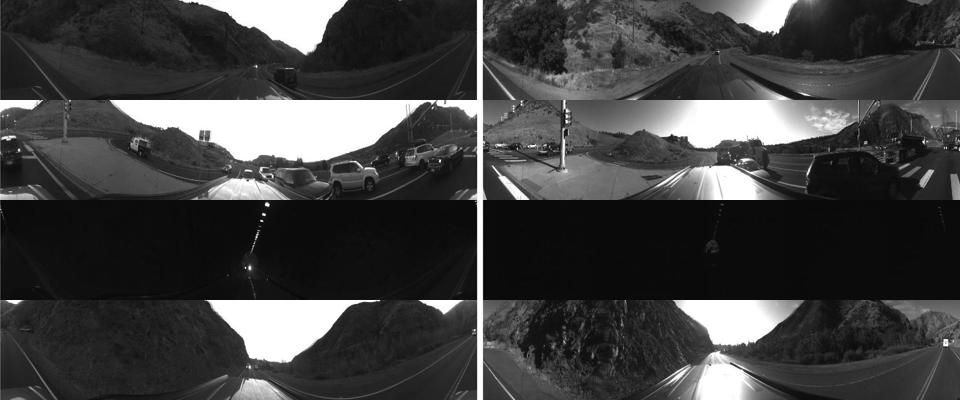}}%
\hfill
\subfloat[Precision-Recall Curves{\label{M_D2}}]{\includegraphics[width=.5\textwidth]{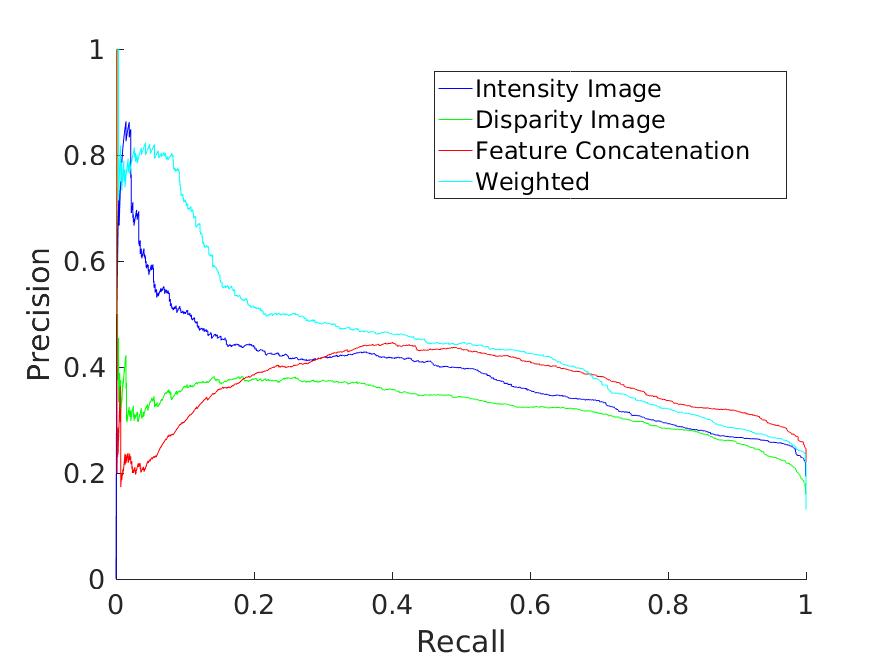}}%
\hfill
\subfloat[Modality Importance Weights{\label{M_D3}}]{\includegraphics[width=.5\textwidth]{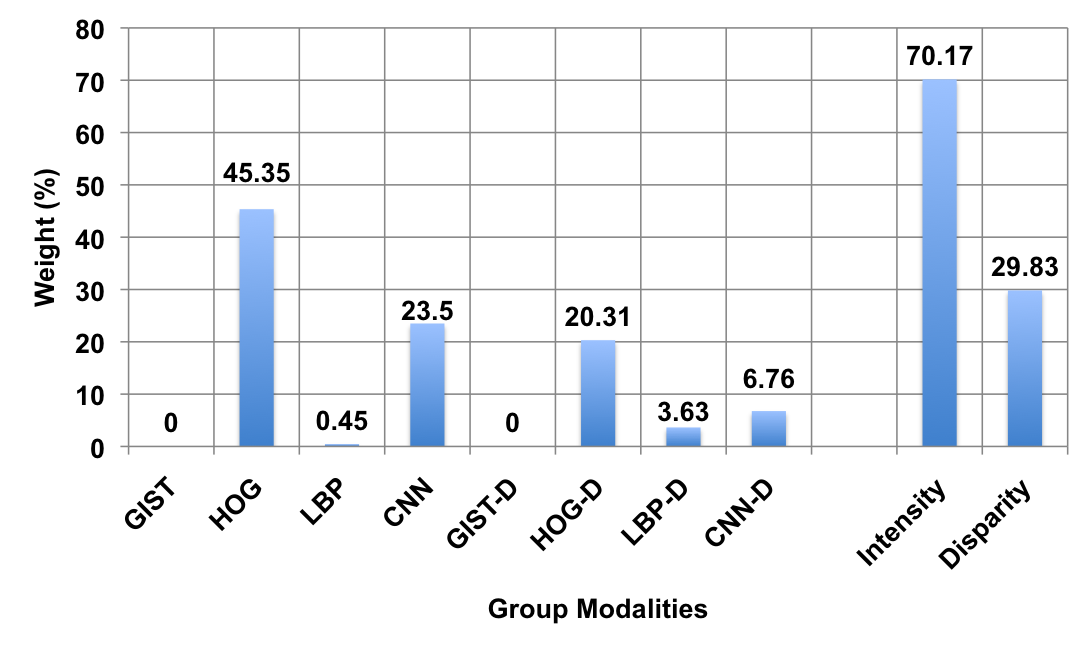}}%
\hfill
\caption{Mountain route -- intensity versus disparity data. Fig \ref{M_D1} shows Fall evening and Summer morning scene matches on the left and right side respectively. These matches were chosen to delineate perceptual aliasing challenges and large illumination changes on the mountain route. Fig. \ref{M_D2} illustrates a performance comparison using precision-recall curves. The weighted line represents the results from our optimization algorithm. Fig \ref{M_D3} illustrates modality importance weights. Each feature modality was extracted from each sensor modality. Feature modalities from disparity images have an additional '-D' at the end.}\label{fig:mountd}
\end{figure*}

\subsection{Effectiveness of Disparity Modality for Long-term Place Recognition}
\label{sec:DI}
Two distinct outdoor environments (city and mountain routes) were studied to analyze how disparity images can aid intensity image for loop closure detection. 
For both the routes, summer morning and fall evening scenarios are used during the training phase.
For the city route, frames 400-800 are used for training and frames 1-399 are used for testing. 
Whereas for the mountain route, frames 200-599 are used for training and frames 750-849 are used for testing. 
Only the forward direction was used for this experiment. City and mountain route results are presented in Figure \ref{fig:cityd} and Figure \ref{fig:mountd} respectively.

As seen in Figure \ref{C_D3} and \ref{M_D3}, our analysis indicates that the intensity image is the dominant sensor modality for the city and mountain route with higher weight percentages. Although, it is important to point out that the disparity image has a larger importance weight in the city (more feature abundant environment) than the mountain route. Hence, the disparity image could aid in capturing discriminative scene signatures in environments similar to the city route, where the geometry of the environment is more defined. Furthermore, the disparity image can also be utilized to recognize discriminative geometrical patterns in the scene, for short term place recognition. Our analysis further indicates that the accuracy of the disparity image is not as good as the intensity image. This can be observed in Figure \ref{C_D2} and \ref{M_D2}. Our optimization algorithm takes into account the discrepancies in the importance weight between the two sensor modalities, yielding in better precision and accuracy than conventional feature concatenation methods. Choosing only the highest weighted modalities can allow users to take advantage of multiple sensors without putting excessive burden on the onboard processors.

\subsection{Bidirectional Loop Closure Detection}
\label{OLP}
Both the city and mountain route datasets were utilized to assess bidirectional loop closure detection. Figure \ref{fig:route} illustrates how we defined forward and backward direction for the city and mountain route.  We train the model using the forward summer morning and forward fall evening scenarios and then test for loop closure detection on the backward fall evening and backward summer evening for the city and mountain routes respectively. The same frames used in section \ref{sec:DI} for training are used for both routes in this experiment. From backward datasets, frames 2262-2662 are used for the city route and frames 4570-4970 are used for the mountain route.

\begin{figure*}[t!]
\centering
\hspace*{\fill}
\subfloat[Mountain Route Scene Match{\label{R_A}}]{\includegraphics[width=1\textwidth]{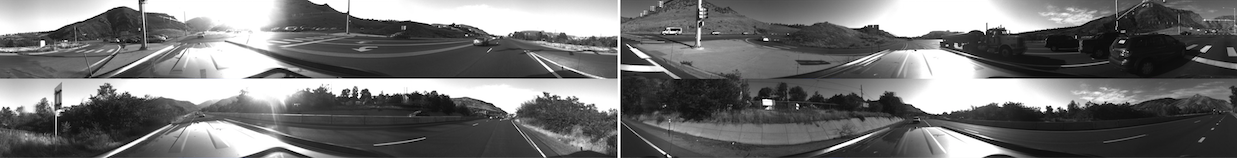}}%
\hfill
\subfloat[City Route Scene Match{\label{R_B}}]{\includegraphics[width=1\textwidth]{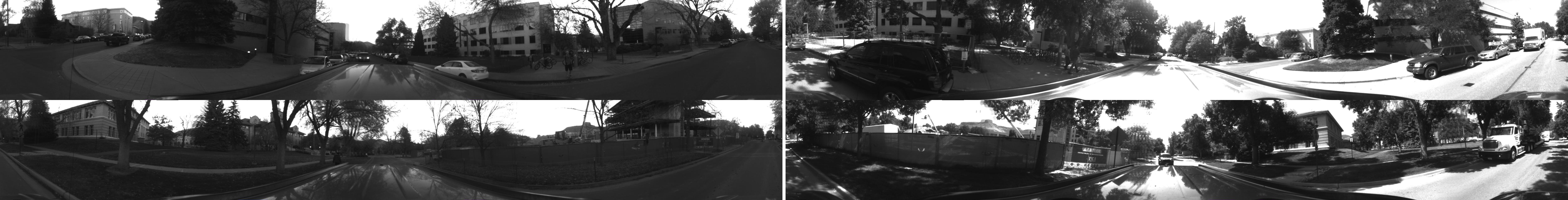}}%
\hfill
\subfloat[Precision-Recall Curve (Mountain Route){\label{R_C}}]{\includegraphics[width=.5\textwidth]{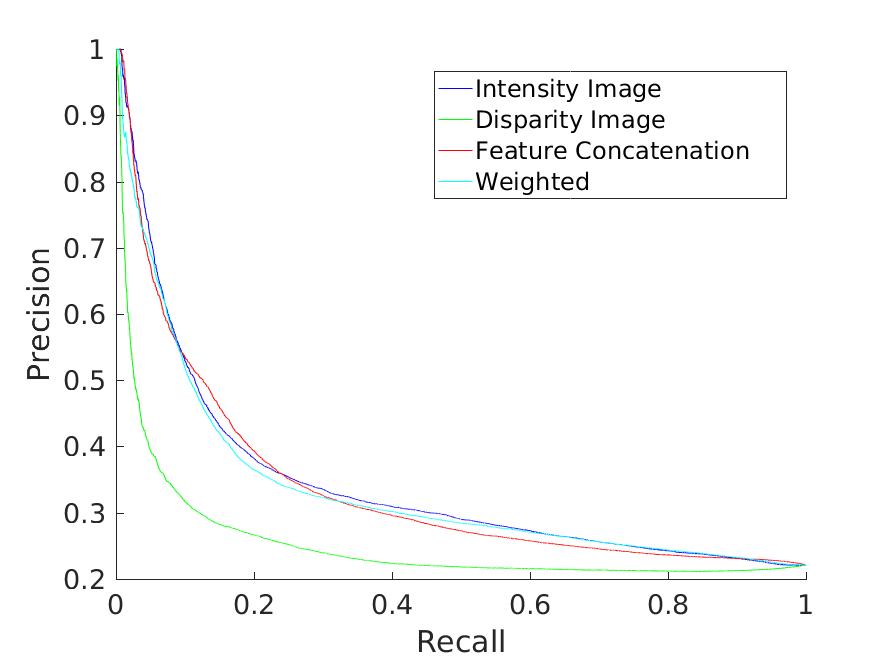}}%
\hfill
\subfloat[Precision-Recall Curve (City Route){\label{R_D}}]{\includegraphics[width=.5\textwidth]{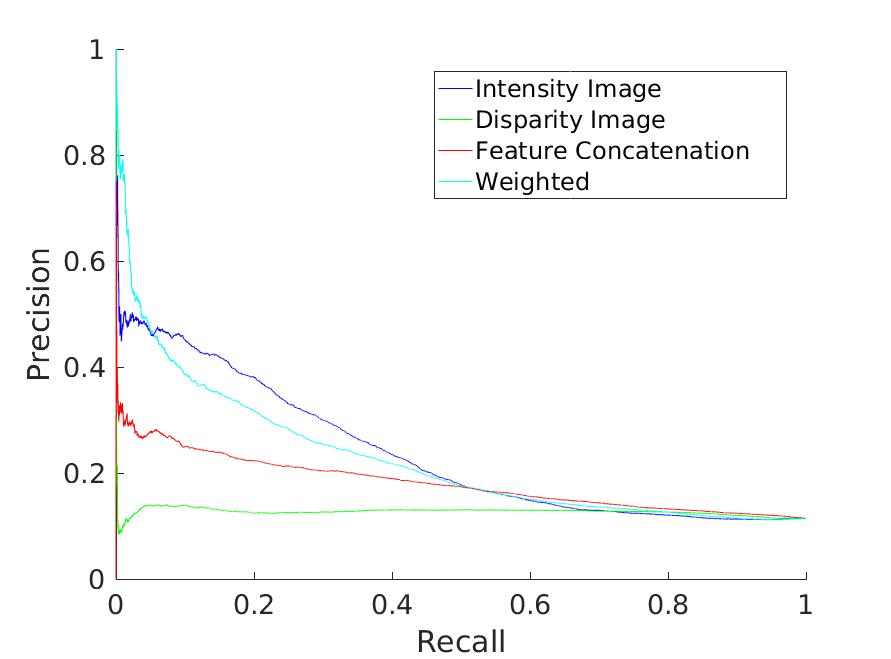}}%
\hfill
\caption{Mountain and city route -- bidirectional loop closure detection. Fig (\ref{R_A})(\ref{R_B}) presents backward - Fall evening and forward - Summer morning scene matches on the left and right side respectively. Fig. (\ref{R_C})(\ref{R_D}) illustrates a performance comparison using precision-recall curves. The weighted line represents the results from our optimization algorithm.}\label{fig:R}
\end{figure*}

As it can be observed from the scene matches in Figure \ref{R_A} and \ref{R_B}, despite extreme illumination changes, dynamic activities (e.g. construction) and occlusions from trees and other vehicles, successful loop closures were detected. Due to the omnidirectional vision, the required combination of discriminative features can be captured. In the presence of all the previously described challenges, using a monocular camera might not yield the required discriminative features, as they might be outside the field of view. The accuracy of our optimization algorithm is presented in Figure \ref{R_C} and \ref{R_D}. For both the routes, our sparsity inducing algorithm achieves similar or better results than simple feature concatenation, demonstrating the importance of capturing discriminative scene signatures for successful bi-directional loop closure detection.

\section{Conclusion and Future Work}
\label{sec:con}
In this paper, we have presented the new MOLP dataset comprising omnidirectional intensity and disparity information of two very distinct routes. 
The dataset presents a wide range of outdoor characteristics including long-term appearance variations due to dynamic activities (e.g., traffic, pedestrians), illumination and weather variations, 
traditional vision challenges (e.g., camera blurs) and perceptual aliasing challenges.
Our MOLP dataset is a comprehensive benchmark, which is released as open source. 
In addition to this, we have shown that the intensity image is able to better capture long term appearance changes as compared to disparity images. 
However, the importance weight of the disparity image is still significant for both the city and mountain routes. Furthermore, we have also presented promising results using omnidirectional vision along with our sparse convex optimization algorithm for bidirectional loop closure detection. By understanding the importance weight of the sensor and feature modalities used, one can omit the information provided by low weight sensor and feature modalities, easing computational burden on the robot, while still being able to augment the system with multiple sensor and feature modalities for robust results. Possible future work involves developing an online place recognition system which uses this discriminative feature isolating property for efficient and accurate loop closure detection. Furthermore, research into using only a subset or important camera view angles can further reduce the burden on onboard processors.

\bibliographystyle{../styles/spbasic}
\bibliography{../MOLP}

\end{document}